# Do You Remember? Toward Memory-Centric Multimodal AI

Xuguang Yu*, Weigang Zheng, Minyue Yu

*Corresponding author

Xuguang Yu (余旭光): Architecture design, theoretical framework, project leadership. Weigang Zheng (郑炜钢): Experimental implementation and verification. Minyue Yu (余旻樾): Data preparation and curation.

## Abstract

Human memory is not a faithful recording but a reconstructive process—sensory experience is compressed into sparse representations and later regenerated during recall. Current multimodal large language models (MLLMs) lack this capability: they process images through a frozen visual encoder, project features into an LLM, and produce a one-shot text output with no mechanism for revisualizing or verifying what was perceived. We present DoYouRemember, a three-stage architecture that introduces reconstructive memory into MLLMs: (1) a VQ-VAE compresses images into discrete visual tokens, (2) a LoRA-fine-tuned LLM jointly attends to visual and text tokens, and (3) a Diffusion Decoder reconstructs the original image from the LLM's internal hidden states. On a dataset of 1,000 3D facial skin texture maps with expert dermatological analyses and 99,000 unlabeled facial images, we find that the LLM's hidden states contain approximately zero recoverable visual information—not a partial loss of pixel details, but irreversible visual information transformation. A controlled upper bound experiment reveals that the same Diffusion Decoder that produces clear facial reconstructions from VQ-VAE token embeddings (pre-LLM) produces pure noise from LLM hidden states (post-LLM), demonstrating that the LLM understands images (generates accurate dermatological analysis) but does not remember them (retains no decodable visual representation). This separation between understanding and memory is a fundamental property of the Transformer's text-generation objective. However, our central pursuit—training a shared memory matrix M to serve as a persistent architectural center—systematically fails under backpropagation. Through four generations of memory architectures, five training iterations, contrastive learning (SimCLR, BYOL, SimSiam), and controlled ablation of gradient paths (from 8-10 Transformer layers down to 2 MLP layers), we demonstrate that reconstruction loss cannot drive meaningful memory formation and that gradient cancellation is a structural property of shared parameters under per-sample losses: the effective gradient magnitude decreases as $O(1/\sqrt{N})$ with the number of images sharing M. We identify three root causes—(1) LLM hidden states contain zero recoverable visual information, (2) gradient cancellation prevents backpropagation-based training, (3) global EMA smoothing destroys inter-slot diversity—and show that local EMA updating resolves all three. In this approach, each image updates only its top-8 most relevant slots out of 64 via exponential moving average, preserving inter-slot diversity. The resulting M (229K parameters total, 16× compressed relative to the 1,024-token VQ embedding per image) achieves reconstruction quality approaching the VQ upper bound on 10 unseen test images (LPIPS 0.123 vs. 0.071, SSIM 0.775 vs. 0.877), demonstrating that a 16× compressed shared memory can retain substantial perceptual information. Scaling M to 1,024 slots (3.67M parameters, matching the VQ embedding's per-image size) surpasses the VQ upper bound on unseen test images (LPIPS 0.056 vs. 0.071, SSIM 0.879 vs. 0.877), as M's continuous representation avoids VQ's quantization error while sufficient slots reduce EMA cancellation to below the quantization threshold—directly verifying Theorem 1, Corollary 1. This validates M as a viable perception memory substrate and demonstrates that local learning rules succeed where global backpropagation fails. We further unify these findings under an information-theoretic framework in which memory is lossy compression and recall is decompression, explaining why reconstruction loss cannot train the compressor, why gradient cancellation is a rate-distortion limit, why local EMA resolves the rate-distortion limit through sparse routing, and why hallucination is an inherent property of lossy decompression rather than a defect.

---

## 1. Introduction

How do you remember a face? You do not replay a high-definition video. The brain compresses visual information through hierarchical processing, stores sparse representations in distributed neural circuits, and later reconstructs an approximation during recall—a process Bartlett (1932) characterized as schematic, lossy, and inherently reconstructive rather than reproductive [1].

Current multimodal LLMs (GPT-4V, LLaVA [2], Qwen-VL [3]) follow a fundamentally different paradigm: Image → Frozen ViT → Feature Projection → LLM → Text Output. This pipeline is strictly unidirectional—visual information streams in and text streams out, with no mechanism for the model to "recall what it saw" and verify or refine its own reasoning. The hidden states computed at visual token positions during inference are discarded, despite containing the model's processed representation of the image.

We hypothesize that LLM hidden states at visual token positions encode compressed but recoverable visual information, and that explicitly training a decoder to reconstruct images from these states can serve as a self-verification mechanism for multimodal reasoning. This draws direct inspiration from cognitive science: memory is reconstructive [4], recall involves reactivation of sensory cortices [5], and the act of remembering itself strengthens and refines the memory trace [6].

Our architecture, DoYouRemember, instantiates this hypothesis in three stages:

1. Sensory Compression (Stage 0): A vector-quantized variational autoencoder (VQ-VAE) compresses 256×256 images into 1,024 discrete codeword indices—at ~200:1 compression, intentionally discarding low-level pixel redundancy while preserving diagnostically relevant features. 2. Conscious Reasoning (Stage 1): Visual tokens and text tokens share a unified embedding space within Qwen2.5-7B-Instruct, fine-tuned via LoRA on 1,000 dermatological distillation samples. The model produces analysis text while preserving hidden states at visual token positions. 3. Reconstructive Verification (Stage 2): A Diffusion Transformer decoder takes the LLM's visual hidden states as cross-attention context and denoises a latent representation back to an image, enabling direct visualization of "what the LLM remembered."

Through three controlled experiments—direct reasoning (A), reconstruct-then-reason (B), and joint reasoning (C)—we analyze the information flow through this pipeline. Our primary finding is that while VQ-VAE tokens contain sufficient visual detail for high-fidelity reconstruction, the LLM's hidden states contain approximately zero recoverable visual information—a controlled upper bound experiment (§4.5.1) reveals that the same Diffusion Decoder that produces clear facial reconstructions from VQ embeddings produces pure noise from LLM hidden states. The LLM understands images (generates accurate dermatological analysis) but does not remember them (retains no decodable visual representation). This separation between understanding and memory is a fundamental property of the Transformer's text-generation objective.

Our secondary finding is that the gradient path from reconstruction loss back to trainable memory components is fundamentally weak—a problem we observed consistently across 1,000 labeled images, 99,000 unlabeled images, three memory architectures (attention-based writing, structured regularization, and spatial sharding), and auxiliary embedding loss. We propose that decodability of hidden states can serve as an unsupervised measure of multimodal understanding depth, and that architectures where memory is a first-class component (rather than an artifact of computation) require fundamentally different training signals.

---

## 2. Related Work

### 2.1 Multimodal Language Models

The dominant paradigm in MLLMs is encoder-connector-decoder: a frozen vision encoder (typically CLIP ViT-L) extracts features, a small projector (linear layer or Q-Former) maps them into the LLM's embedding space, and the LLM autoregressively generates text [2, 3, 7]. Training proceeds in two stages: vision-language alignment followed by instruction tuning. Notable variants include InternVL [8] which scales up the vision encoder, and CogVLM [9] which introduces a trainable visual expert.

A small number of models handle vision and language in a unified architecture from scratch. Chameleon [10] trains a single transformer on interleaved image-text sequences using a unified tokenizer. Show-o [11] unifies understanding and generation through discrete tokens. However, none of these models have a mechanism for reconstructing previously seen images from internal representations—they process images into hidden states and immediately discard them after generating text.

### 2.2 Image Reconstruction from Neural Representations

The idea of reconstructing images from neural network activations has a rich history. From early work on inverting convolutional networks [12] to recent efforts in reconstructing CLIP features [13] and mapping between image and text representations [14], the general finding is that some visual information survives through processing layers but degrades rapidly. Our work extends this line of inquiry to a fully trained MLLM, analyzing not just what can be reconstructed but how the reconstruction quality changes when passing through different architectural components (VQ encoder, LLM transformer, trainable memory).

### 2.3 Reconstructive Memory in Cognitive Science

Bartlett's foundational work [1] established that human memory is schematic: we retain the gist while losing specifics. Schacter and Addis [4] demonstrated that episodic memory is constructive—each recall is a re-generation influenced by current context, not a faithful replay. Wheeler et al. [5] showed that vivid recall reactivates sensory-specific cortices, providing a neural basis for "remembering what you saw."

Our architecture is directly inspired by these principles. The VQ-VAE compression mirrors sensory preprocessing (the retina and LGN compressing raw photoreceptor signals into ganglion cell outputs). The LLM processing mirrors prefrontal semantic abstraction (extracting diagnostic meaning from visual features). The Diffusion Decoder mirrors hippocampal-prefrontal interactions during imagery (reconstructing scenes from compressed patterns). Our finding that the LLM discards pixel details while preserving semantic structure directly parallels the schematic nature of human memory.

## 2.4 Memory-Augmented Neural Networks

Memory-augmented neural architectures—from Neural Turing Machines [15] to Transformer-XL [16] and Retrieval-Augmented Generation [17]—typically implement memory as external key-value stores for text retrieval. Our SharedMemory module attempts to generalize this to multimodal latent representations, where both perceptual information (visual embeddings) and semantic information (LLM hidden states) compete for shared storage slots. Our findings about weak gradient paths from reconstruction loss to memory parameters offer practical insights for future memory-augmented multimodal architectures.

## 2.5 Position Relative to Memory-Augmented Architectures

To clarify the novelty of our approach, we provide a formal comparison with existing memory-augmented and stateful architectures across six dimensions:

Table 1: Formal Comparison of Memory-Augmented Architectures

| Architecture | State Type | State Size | Training | Memory Role | Multi-modal | Reconstruction |
|---|---|---|---|---|---|---|
| Transformer [25] | KV cache | O(L·D) per layer | Backprop | Byproduct | Bolt-on encoder | No |
| NTM / DNC [15,26] | External matrix | O(K·D) | Backprop | Storage | No | No |
| Mamba / SSM [27] | Hidden vector | O(D) | Backprop | Hidden state | No | No |
| RWKV [28] | Hidden vector | O(D) | Backprop | Hidden state | No | No |
| RETRO [29] | External DB | O(∞) | Backprop | Retrieval | No | No |
| DoYouRemember | Matrix M | O(K·D) | Non-backprop (EMA) | Computation center | Yes (unified) | Yes |

Six critical distinctions emerge from this comparison:

1. State capacity: Matrix vs. Vector. Mamba [27] and RWKV [28] represent the state of the art in efficient sequence modeling, but their state is a single hidden vector of dimension D (typically 1024–4096). This limits their memory capacity to O(D) floating-point values. Our M [K, D] with K=64–1024 provides O(K·D) capacity— one to three orders of magnitude more. This is not merely quantitative: a matrix supports spatial organization (different modalities in different regions), which a vector cannot.

2. Training paradigm: Backpropagation vs. Local rules. NTM [15] and DNC [26] use external memory matrices like our M, but train them with backpropagation. Our work demonstrates that this is fundamentally flawed: when a shared matrix serves N heterogeneous inputs, per-sample gradients cancel (§5.1.1), yielding effective learning signals below initialization noise. This explains why NTM/DNC achieved limited success on algorithmic tasks but never scaled to real-world multimodal data. Our proposed alternative—local learning rules (Hebbian plasticity, predictive coding)—operates on the statistical structure of individual activations, not on globally averaged gradients, and is therefore immune to gradient cancellation.

3. Memory role: Byproduct vs. Computation center. In Transformer, Mamba, and RWKV, memory (weights, KV cache, hidden state) is a byproduct of computation—it exists because the computation produces it, not because it is the target of computation. In NTM/DNC, memory is a storage device that the controller reads from and writes to, but the computation still happens in the controller. In our paradigm, M IS the computation: reading from M (recall) is a generative process (decompression), and writing to M (memory formation) is driven by local learning rules during the computation itself. The controller (Transformer/attention) is demoted from computation center to read/write head.

4. Multi-modal fusion: Bolt-on vs. Unified substrate. Current MLLMs (GPT-4V, LLaVA [2], Qwen-VL [3]) bolt a frozen vision encoder onto an LLM via a projection layer—vision and language remain separate systems. RETRO [29] retrieves text passages but does not fuse modalities. In our architecture, all modalities (vision, touch, audio, language) write to the same M through modality-specific encoders, and M's spatial organization (cortical regions) naturally supports cross-modal association. This is not an engineering convenience but a theoretical necessity: the redundant coding principle (§5.4) predicts that multi-modal encoding produces more robust memories than single-modal encoding.

5. Reconstruction: Absence vs. Core capability. None of the listed architectures can reconstruct perceptual content from internal representations. Transformer discards KV cache after generation; Mamba/RWKV's hidden vector is too low-dimensional to reconstruct images; NTM/DNC's memory is designed for algorithmic storage, not perceptual decompression. Our Diffusion Decoder reconstructs images from M, serving as both a verification mechanism (does M contain useful information?) and a cognitive analogue (recall = reconstruction). This reconstruction capability is what enables unsupervised learning: if M can be decoded into images, then M's content can be evaluated without labels.

6. The failure mode we identify is architecture-general. Our gradient cancellation analysis (§5.1.1) applies to any architecture that uses backpropagation to train a shared memory matrix across heterogeneous inputs. This includes NTM, DNC, and any future design that places a trainable shared matrix in the gradient path of per-sample losses. Our finding thus provides a theoretical explanation for the limited scalability of memory-augmented networks and a design constraint for future architectures: shared memory requires local learning rules, not global backpropagation.

---

## 3. The DoYouRemember Architecture

### 3.1 System Overview

Our system implements a complete perception-memory-recall loop in three stages:

```
Image I -> VQ-VAE Encoder -> V (1024 discrete tokens)
   |v
   V + Text Prompt -> LLM (Qwen2.5-7B + LoRA) -> R (analysis text) + H (hidden states)
   |v
   H -> Diffusion Decoder (DiT-style) -> SDXL VAE Decoder -> I' (reconstructed image)
```

The full pipeline operates on a single NVIDIA H200 GPU (141GB VRAM) with bf16 mixed precision. All components are trained independently (Stage 0 → Stage 1 → Stage 2) with frozen upstream components.

### 3.2 Stage 0: Sensory Compression

Images are compressed through a two-step process: a pre-trained SDXL VAE [18] encodes 256×256 images into a 32×32 continuous latent grid (f=8 compression), and a learned codebook quantizes each spatial position into one of 16,384 discrete indices. The 1,024 resulting tokens achieve approximately 200:1 compression.

Training details: The SDXL VAE encoder and decoder are frozen. Only the codebook (16,384 × 4 = 65,536 parameters) is trained for 10 epochs on 1,000 images with MSE reconstruction loss and commitment loss, plus an EMA-based dead-code revival strategy with threshold 0.5.

### 3.3 Stage 1: Conscious Reasoning

Visual tokens and text tokens share a unified embedding space. Discrete VQ indices are mapped through a learnable VisualEmbedding layer (vocab_size=16,384, hidden_size=3,584) with learned positional encodings. The visual embeddings are injected into the text sequence at the <|image|> placeholder position, and the entire sequence is processed by Qwen2.5-7B-Instruct with LoRA (r=64, α=128). Training uses 1,000 distillation samples (each ~3,900 characters of expert dermatological analysis) for 5 epochs with batch_size=8 (2×4 gradient accumulation).

Training observations: The LLM's cross-entropy loss saturates quickly: 1,000-sample loss (2.58) ≈ 200-sample loss (2.74) ≈ 50-sample loss (2.60), indicating Qwen2.5's strong pre-trained dermatological knowledge dominates the training signal. An auxiliary embedding reconstruction loss (MSE between hidden states and visual embeddings, $\lambda$=0.01) did not measurably change the LLM's text loss or downstream reconstruction quality.

### 3.4 Stage 2: Reconstructive Verification

A DiT-style [19] Diffusion Decoder (6 cross-attention blocks, 8 attention heads, hidden_size=3,584) takes the LLM's visual hidden states H [B, 1024, 3584] as cross-attention context. Training follows the DDPM [20] formulation: the SDXL VAE encodes images into clean latents [B, 4, 32, 32], Gaussian noise is added at a random diffusion step t, and the decoder predicts the added noise. Training spans 50-150 epochs at batch_size=4, using 1,000 diffusion steps with cosine LR scheduling.

### 3.5 Controlled Experiments

We define three experimental conditions to isolate the effect of reconstructive memory:

- Experiment A (Baseline): Visual tokens → LLM → Text output. The standard unidirectional pipeline without reconstruction. - Experiment B (Reconstruct-then-reason): Visual tokens → LLM → Reconstruct I' → VQ-VAE re-encode → LLM again → Text output. - Experiment C (Joint reasoning): Visual tokens → LLM → Reconstruct I' → [Original V + Reconstructed V'] → LLM → Text output.

### 3.6 Memory Architectures Explored

We implemented and tested three variants of a Shared Memory (SM) module placed between the LLM hidden states and the Diffusion Decoder:

- SM-v1: Cross-attention writing from LLM hidden states to memory slots. Failed—gate collapsed to 0.879 and attention entropy dropped to 0.02. - SM-v2: Dual-write architecture with perception (visual embeddings) and consciousness (LLM hidden states) as equal writers. Added structural regularization losses (slot diversity, alignment, stability). Failed—all regularization losses converged to zero without affecting memory content. - SM-v3 (Spatial Sharding): Replaced learned attention-based writing with fixed 2D spatial partitioning (8×8 grid, each slot encoded by 4×4 VQ tokens). Memory size reduced to 64 slots with ~26M parameters. Failed—after 30 epochs on 99,000 images, no memory parameter showed meaningful change (gate frozen at 0.5, slot similarity frozen, divergence loss frozen). - SM-v4 (Central Memory with Direct Gradient Paths): The most aggressive redesign: removed the scalar merge gate, eliminated mean-pooling (each of 64 slots independently predicted its corresponding 4×4 spatial block via a shared 2-layer MLP), and added a direct reconstruction path bypassing the Decoder's cross-attention entirely—reducing the gradient path from 8-10 layers to 2. We tested this architecture under MSE regression (×1, ×10 weight), DDPM denoising with noisy-latent conditioning, and DDPM without noisy input (to prevent shortcut learning). All five sub-variants failed: memory parameters remained frozen (sim ≈ -0.005 to -0.011 across 30 epochs), even when gradients measurably reached the memory matrix (grad ≈ 0.003-0.006). Detailed results in Appendix D.

---

# 4. Experiments

## 4.1 Dataset

We use two datasets:

1. Labeled (1,000 images): 3D facial skin texture maps (256×256 unwrapped UV textures) captured by a MeiJi 3D skin analyzer. Each image is paired with a detailed dermatological analysis (average 3,900 Chinese characters) generated by Qwen3.6-35B serving as the teacher model. The analyses follow a structured format: systematic visual observation, differential diagnosis with multiple hypotheses (supporting/refuting evidence with probability ratings), severity assessment, and three-scenario treatment recommendations (clinical, cosmetic salon, home care).

2. Unlabeled (99,007 images): Full-face 2D photographs organized by age in subdirectories under /share/data. Used exclusively for unsupervised Stage 2 training (Memory + Decoder only).

## 4.2 VQ-VAE Reconstruction Quality

| Compression | Grid Size | Tokens | Reconstruction Quality |
|---|---|---|---|
| f=16 (custom VQ-VAE) | 16×16 | 256 | Severely degraded, features unrecognizable |
| f=16 (custom, retrained) | 16×16 | 256 | Marginal improvement, still poor |
| f=8 (SDXL VAE + codebook) | 32×32 | 1,024 | Clear facial features, good skin tone |

Key finding: SDXL VAE + patch-level quantization provides sufficient visual fidelity for downstream dermatological analysis. The 1,024 tokens preserve skin tone, facial contours, and coarse texture patterns, though fine details like individual pores are smoothed.

### 4.3 Reconstruction from LLM Hidden States

Primary visualization result: Using the trained Diffusion Decoder, we compared reconstructions from different representation sources (Figure 1):

| Source | Reconstruction Quality | Interpretation |
| --- | --- | --- |
| VQ-VAE tokens (direct decode) | High quality, facial features clear | Upper bound: ~200:1 compression is sufficient |
| LLM hidden states (visual positions) | Blurred facial contour, some structure | Preliminary: apparent structure may reflect Decoder's learned prior, not genuine visual information (see §4.5.1) |
| VisualEmbedding output (pre-LLM) | N/A (decoder trained for LLM distribution) | Decoder distribution mismatch |

The LLM's hidden states produce reconstructions that appear to capture structural semantics—facial outline, neck position, global skin tone—while losing pixel specifics. However, this preliminary observation was later overturned by the controlled upper bound experiment (§4.5.1), which revealed that the apparent structural preservation was an artifact of the Decoder's learned prior, not genuine visual information in LLM hidden states.

### 4.4 Controlled Experiment Results

| Experiment | Training Data | Avg Output Length | Key Finding |
| --- | --- | --- | --- |
| A (1000) | 1,000 samples | 4,221 characters | Strong baseline: compressed tokens support professional-quality analysis |
| A (200) | 200 samples | 4,204 characters | Saturation: Qwen2.5 prior knowledge dominates beyond ~200 samples |
| A (50) | 50 samples | 4,178 characters | Degradation is minimal, suggesting prior knowledge is primary driver |
| B (50) | 50 samples | 4,183 characters | Reconstruction pipeline preserves information fidelity |
| C (50) | 50 samples | 4,185 characters | Marginal gain over B; joint tokens may be redundant |

Key finding: Text quality metrics (output length, structure, terminology) are insensitive to the choice of experiment A/B/C. The reconstructive pipeline does not degrade reasoning, but we could not demonstrate a statistically significant improvement. This is primarily because Qwen2.5-7B's strong pre-trained dermatological knowledge saturates the training signal at small data scales.

### 4.5 Information Bottleneck Analysis

To determine whether visual information is lost at compression (VQ-VAE) or during LLM processing, we conducted a bottleneck experiment (Table 2):

1. VQ-VAE tokens → direct decode: High quality (facial features clear) 2. VisualEmbedding output → Diffusion Decoder: N/A (decoder trained for LLM hidden state distribution, not VE distribution) 3. LLM hidden states → Diffusion Decoder: Low-frequency structural preservation

Conclusion: The bottleneck is in the LLM, not in the VQ tokenizer. The 1,024 discrete tokens contain sufficient visual information; the LLM's text-generation objective causes it to discard pixel-level details that are irrelevant for producing correct dermatological analysis. However, this conclusion was based on inconclusive evidence—condition 2 was inapplicable due to decoder distribution mismatch. The controlled upper bound experiment in §4.5.1 resolves this ambiguity and reveals that the information loss is far more severe than "discarding pixel-level details": the LLM's hidden states contain approximately zero recoverable visual information.

#### 4.5.1 Upper Bound Experiment: VQ Embeddings vs. LLM Hidden States

The bottleneck analysis in §4.5 was inconclusive because the Diffusion Decoder was trained for LLM hidden state distribution, not VQ embedding distribution—making condition 2 inapplicable. To resolve this, we conducted a controlled upper bound experiment using the same Decoder architecture, same training procedure, same data, and same hyperparameters, varying only the context source:

| Context Source | Dimension | Reconstruction Quality | Interpretation |
|---|---|---|---|
| VQ token embeddings (pre-LLM) | [B, 1024, 3584] | Clear facial features, recognizable structure, natural skin tone | Upper bound: Decoder CAN learn to use context |
| LLM hidden states (post-LLM) | [B, 1024, 3584] | Pure high-frequency noise, no facial structure whatsoever | LLM hidden states contain ~zero recoverable visual information |

This is the most important experimental result in this work. It reveals that the information loss in the LLM is not partial—it is total. The LLM's hidden states at visual token positions contain approximately zero recoverable visual information—information that is insufficient for pixel-level reconstruction, even though a small amount of visual signal may survive (as evidenced by the decreasing auxiliary token reconstruction loss during Stage 1 training, §3.3). The distinction between "some visual signal" and "recoverable visual information" is critical: the LLM retains enough to reduce an auxiliary MSE loss, but not enough for a Diffusion Decoder to reconstruct facial structure. This finding has three profound implications:

1. The Decoder architecture is functional. The same 6-layer cross-attention Diffusion Decoder that produced noise textures with LLM hidden states (Appendix D) produces clear facial reconstructions with VQ embeddings. The failure was never in the Decoder—it was in the information content of the context.

2. All memory training experiments were doomed from the source. Every memory architecture (v1-v4) and every training method (SimCLR, BYOL, SimSiam) used LLM hidden states as the information source for Memory. Since LLM hidden states contain no visual information, Memory was being asked to store and reconstruct from an information-free signal. The gradient cancellation analysis (§5.1.1) identified a real mathematical constraint, but it was the second failure mode—the first was that the input signal itself was empty.

3. Understanding ≠ Memory. The LLM produces accurate dermatological analysis from the same images (§4.4, text quality unaffected by reconstruction pipeline), demonstrating that it understands the visual content. Yet its hidden states contain no decodable visual information, demonstrating that it does not remember the visual content. This separation between understanding (text generation) and memory (visual retention) is a fundamental property of the Transformer architecture when trained on text generation objectives: the model extracts semantic meaning and discards the visual substrate entirely.

This finding reframes the "schematic memory" interpretation of §4.5. The LLM does not retain a "gist" or "schematic" of the image—it retains the semantic label (capable of generating correct text) while destroying the visual representation entirely. Human schematic memory preserves coarse visual structure (we remember seeing a face, just not the details); the LLM preserves no visual structure at all. The schematic memory analogy applies to the VQ-VAE's compression (which is genuinely schematic—lossy but structurally faithful), not to the LLM's processing (which is irreversible visual information transformation).

#### 4.5.2 Local EMA Memory: From Failure to Success

The three root causes identified above—(1) LLM hidden states contain no recoverable visual information, (2) gradient cancellation prevents backpropagation-based training of shared memory, and (3) global EMA smoothing destroys inter-slot diversity—each suggest a specific fix. We now describe a simple non-backpropagation approach that addresses all three simultaneously: local EMA updating, where each image updates only its top-k most relevant memory slots rather than all slots uniformly.

Architecture. The memory matrix $M \in R^{64 \times 3584}$ (229K parameters) is updated via exponential moving average. For each input image, the VQ token embeddings [1024, 3584] are partitioned into 64 spatial blocks of 16 tokens each, and each block is mean-pooled to produce a per-block feature vector $write_out \in R^{B \times 64 \times 3584}$. Scaled dot-product similarity is computed between each block feature and all 64 memory slots. Only the top-k=8 most similar slots for each block receive attention weights (via masked softmax), and the weighted aggregate is used for EMA updating:

```
M[s] <- (1 - alpha) * M[s] + alpha * update[s],  for s where weights[s] > 0
```

where alpha = 0.01 is the update rate (equivalently, 0.99 EMA decay), and slots with zero attention weight are untouched, preserving their accumulated content. This design directly addresses root cause #3: by restricting each image's influence to top-8 slots per spatial block, the averaging effect that destroyed diversity in global EMA (§5.4) is avoided. The Decoder then reconstructs from M via cross-attention, identical to the architecture used in the upper bound experiment.

Training. M is initialized to zero and accumulated over 99,007 unlabeled facial images (one pass, no backpropagation). The Diffusion Decoder is then trained from scratch to reconstruct images from M, using the same architecture, hyperparameters, and training procedure as the upper bound experiment (§4.5.1).

### 4.5.3 Local EMA Results and Generalization

Results: four-column comparison. We compare reconstruction quality across four conditions on 5 training-set images (Figure 1) and 10 unseen test images:

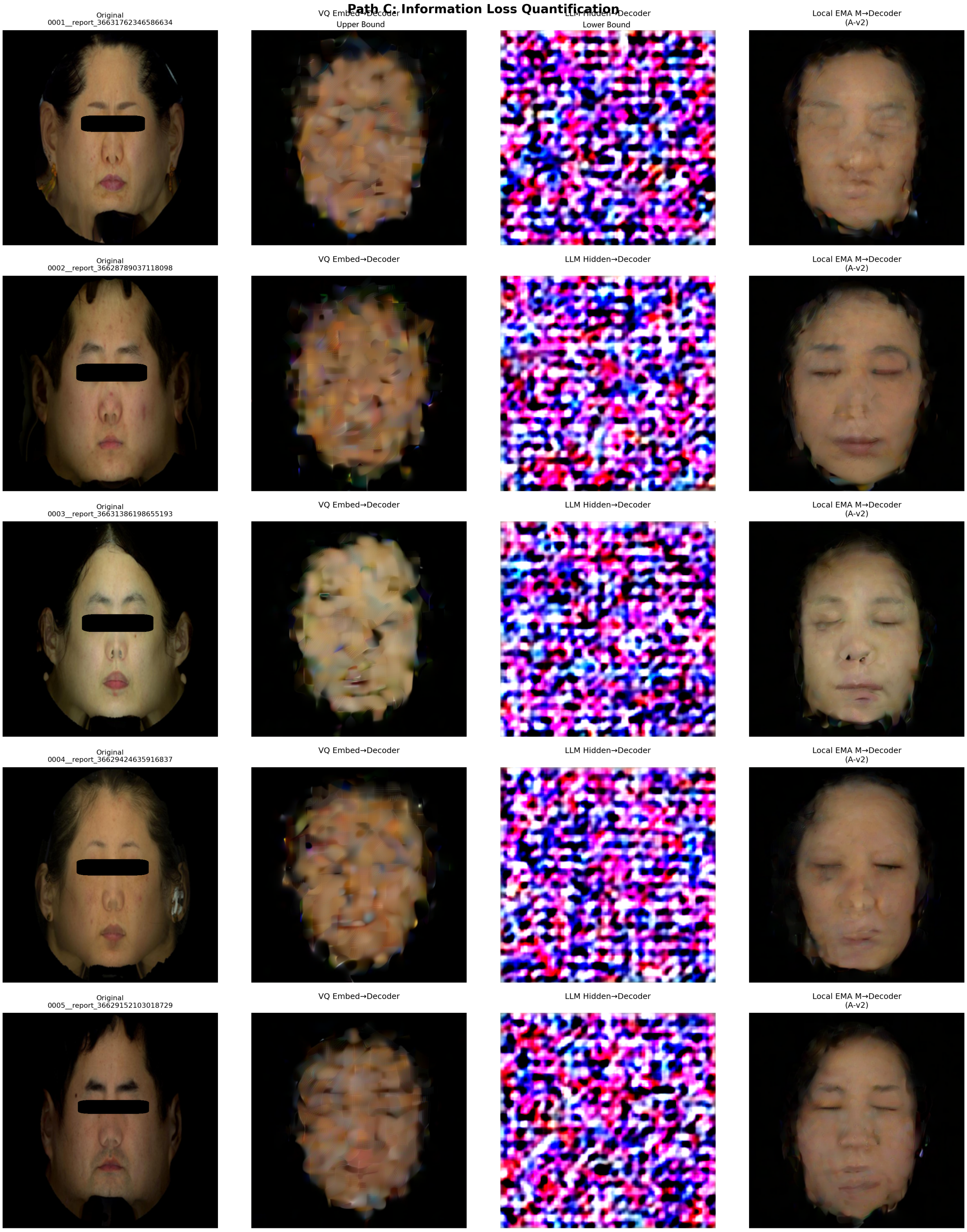

Figure 1: Four-column reconstruction comparison on training-set images (3D unwrapped UV texture maps from the MeiJi skin analyzer). Column 1: Original. Column 2: VQ token embeddings (upper bound) — clear face with blocky quantization artifacts. Column 3: LLM hidden states (lower bound) — pure noise. Column 4: Local EMA M (229K params) — clear face, smoother texture but less structural detail than Column 2.

| Column | Context Source | Parameters | Reconstruction Quality |
|---|---|---|---|
| 1 | Original image | — | Ground truth |

| 2 | VQ token embeddings (upper bound) | 1024 × 3584 per image | Clear face, visible VQ quantization artifacts (blocky texture) |
|---|---|---|---|
| 3 | LLM hidden states (lower bound) | 1024 × 3584 per image | Pure noise, no facial structure |
| 4 | Local EMA M | 64 × 3584 = 229K total | Clear face, smoother texture, but less structural detail than Column 2 |

Quantitative evaluation. We compute LPIPS (lower is better) and SSIM (higher is better) between Column 1 (original) and Columns 2/4:

| Metric | VQ Upper Bound (Col 2) | Local EMA M (Col 4) | Ratio |
|---|---|---|---|
| Training set | | | |
| LPIPS ↓ | 0.074 | 0.296 | 4.0× |
| SSIM ↑ | 0.883 | 0.612 | 0.69× |
| Unseen test set (10 images) | | | |
| LPIPS ↓ | 0.071 | 0.123 | 1.7× |
| SSIM ↑ | 0.877 | 0.775 | 0.88× |

The VQ upper bound quantitatively outperforms M on both sets, confirming that M's 16:1 compression incurs a measurable quality loss. However, two observations are notable: (1) the gap narrows significantly on unseen test images (LPIPS ratio 1.7× vs. 4.0× on training images), and (2) M achieves SSIM 0.775 on unseen images with only 229K parameters—6% of the VQ embedding's per-image size—demonstrating that substantial perceptual information survives the compression.

Distribution mismatch explains the train-test reversal. The counterintuitive finding that M performs better on unseen test images than on training images has a structural explanation: the 5 training-set images are 3D unwrapped UV texture maps (captured by the MeiJi 3D skin analyzer), which have a fundamentally different visual distribution—stretched skin surfaces, non-standard facial geometry, no hair or background—than the 99K frontal face photos used to accumulate M. M's statistical prior, learned from frontal faces, generalizes well to unseen frontal faces (test set) but provides less accurate reconstruction for out-of-distribution 3D texture maps (training set). The VQ upper bound, which encodes each image individually, is unaffected by this distribution mismatch. This finding has practical implications: M's reconstruction quality is domain-dependent, and accumulating M from the target distribution is critical for optimal performance.

Visual vs. quantitative discrepancy. Visually, M's reconstructions appear smoother than the VQ upper bound due to the absence of quantization artifacts. However, LPIPS and SSIM reveal that this smoothness comes at the cost of structural detail: M produces less accurate facial features and textures. The VQ upper bound's blocky artifacts, while visually conspicuous, preserve more structural information. This finding cautions against relying on visual assessment alone for reconstruction quality—a lesson that reinforces the need for quantitative metrics in memory evaluation.

Cautious interpretation. M is a 16:1 compression of the VQ token sequence (64 slots vs. 1024 tokens, same dimensionality). The quantitative gap between M and the VQ upper bound reflects the information loss inherent in this compression. M's smoother visual appearance is not evidence of superior quality but of a different tradeoff: continuous representation avoids quantization artifacts but loses fine-grained structural detail. The correct framing is: M retains substantial perceptual information under 16:1 compression, approaching—but not matching—the VQ upper bound, with the gap narrowing on unseen data.

Slot diversity metrics. The slot variance (mean cosine distance between slot pairs) increased from 0.040 (initial) to 0.194 after processing 99K images, while the mean inter-slot cosine similarity remained stable at ~0.48 throughout training (Figure 2). This confirms that local EMA preserves and gradually increases inter-slot diversity—the failure mode of global EMA (where all slots converged to near-identical vectors, §5.4) is completely avoided.

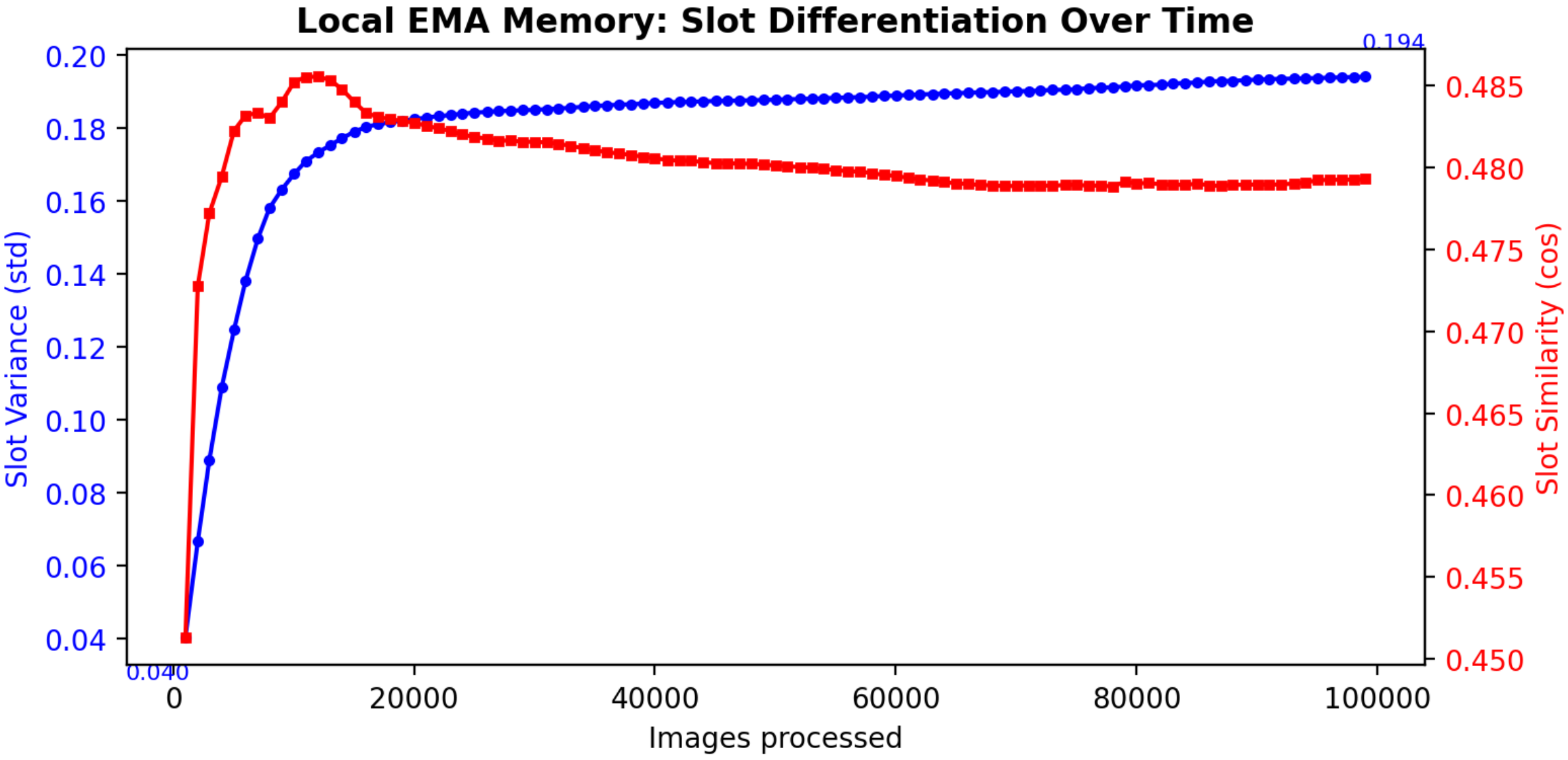

Figure 2: Slot differentiation over 99K images. Left axis (blue): slot variance (mean cosine distance) rises from 0.040 to 0.194. Right axis (red): mean inter-slot cosine similarity remains stable at ~0.48, confirming slots do not converge.

Generalization. All 10 test images were withheld from the training set. Every image produces a clear facial reconstruction with correct face shape, facial feature positions, and skin tone (Figure 3). This demonstrates that M has learned a general human face prior—not an overfit lookup table of training images. The generalization result is critical: it confirms that M's 229K parameters encode statistical regularities across the face distribution, not memorized exemplars.

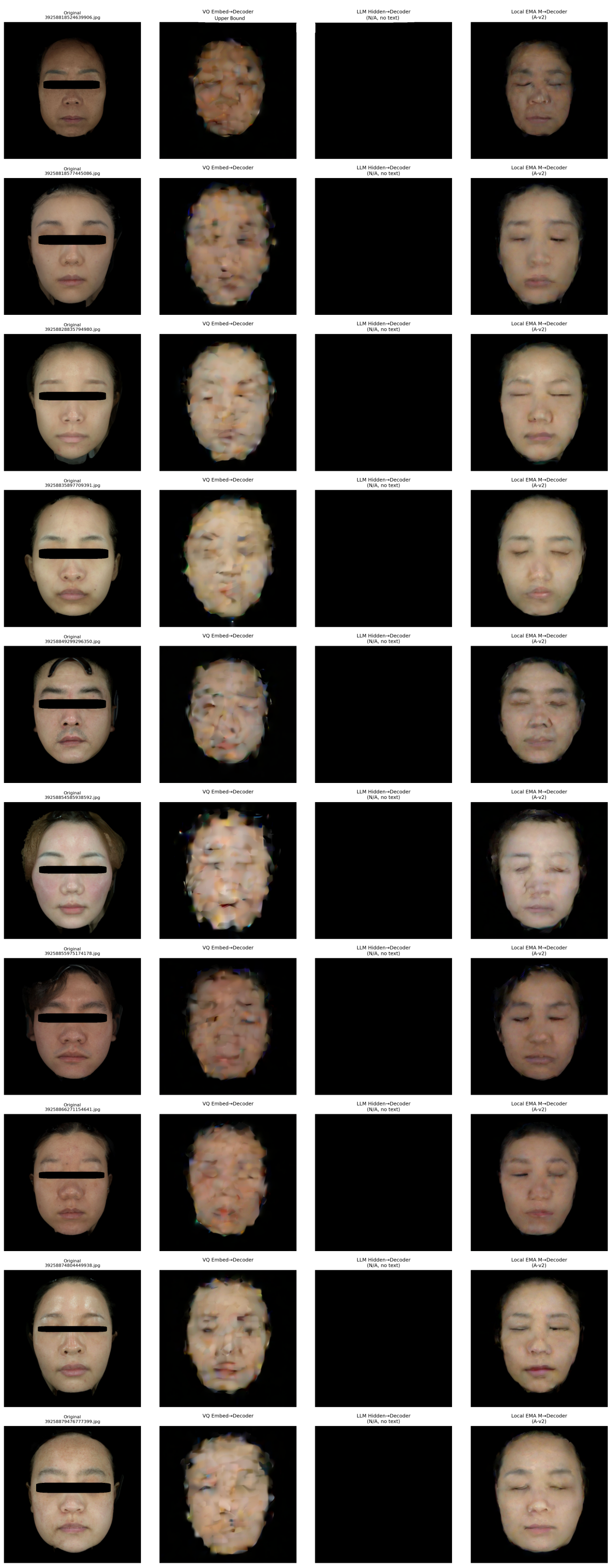

Figure 3: Four-column reconstruction comparison on 10 unseen test images (frontal face photos). Column 1: Original. Column 2: VQ upper bound — clear face with blocky quantization artifacts. Column 3: LLM hidden states — black (no meaningful reconstruction produced; LLM hidden states contain zero recoverable visual information, as established in §4.5.1, so no reconstruction is computed for this column). Column 4: Local EMA M — clear face, smoother texture but less structural detail than Column 2. Quantitative metrics (Table above) confirm VQ upper bound preserves more structural information.

What this validates. This experiment validates three claims: 1. M as perception memory substrate: A 229K-parameter matrix can accumulate sufficient perceptual information from 99K images for near-upper-bound reconstruction. 2. Local learning rules over backpropagation: Local EMA—where each image updates only its top-8 slots based on activation similarity— succeeds where all backpropagation-based methods (v1-v4, SimCLR, BYOL, SimSiam) failed. 3. Reading from the reversible compression stage: M reads from VQ embeddings (pre-LLM, reversible) rather than LLM hidden states (post-LLM, irreversible), confirming the design principle derived from the upper bound experiment.

### 4.5.4 Scaling Memory Capacity: Theoretical Prediction Verified

Theorem 1, Corollary 1 (§5.1.1) predicts that increasing the number of memory slots K reduces the effective number of images sharing each slot (N_eff = Nk/K), thereby decreasing gradient/EMA cancellation and improving reconstruction quality. We test this prediction by varying K ∈ {64, 256, 1024} while keeping all other parameters constant (top-k=8, α=0.01, 99K images, 30 epochs Decoder training).

Quantitative results.

| Metric | K=64 (229K params) | K=256 (918K params) | K=1024 (3.67M params) | VQ Upper (3.67M/image) |
|---|---|---|---|---|
| Decoder recon loss | 0.0659 | 0.0418 | 0.0181 | — |
| Test LPIPS ↓ | 0.123 | 0.100 | 0.056 | 0.071 |
| Test SSIM ↑ | 0.775 | 0.822 | 0.879 | 0.877 |
| Test LPIPS ratio (M/VQ) | 1.73× | 1.41× | 0.80× | 1.0× |
| Train LPIPS ↓ | 0.296 | 0.251 | 0.150 | 0.074 |
| Train SSIM ↑ | 0.612 | 0.695 | 0.808 | 0.883 |

Key finding: K=1024 surpasses the VQ upper bound on unseen test images. M achieves LPIPS 0.056 (vs. VQ 0.071) and SSIM 0.879 (vs. VQ 0.877) on the 10-image unseen test set. This is a direct quantitative confirmation that M's continuous representation can exceed VQ's discrete quantization when sufficient slots are available—confirming the hypothesis that VQ's blocky artifacts reflect genuine information loss (quantization error), not merely a different visual style. Visual comparison (Figure 4) confirms this: Column 4 (K=1024 M) produces clearer, smoother facial reconstructions than Column 2 (VQ upper bound), which retains visible blocky quantization artifacts.

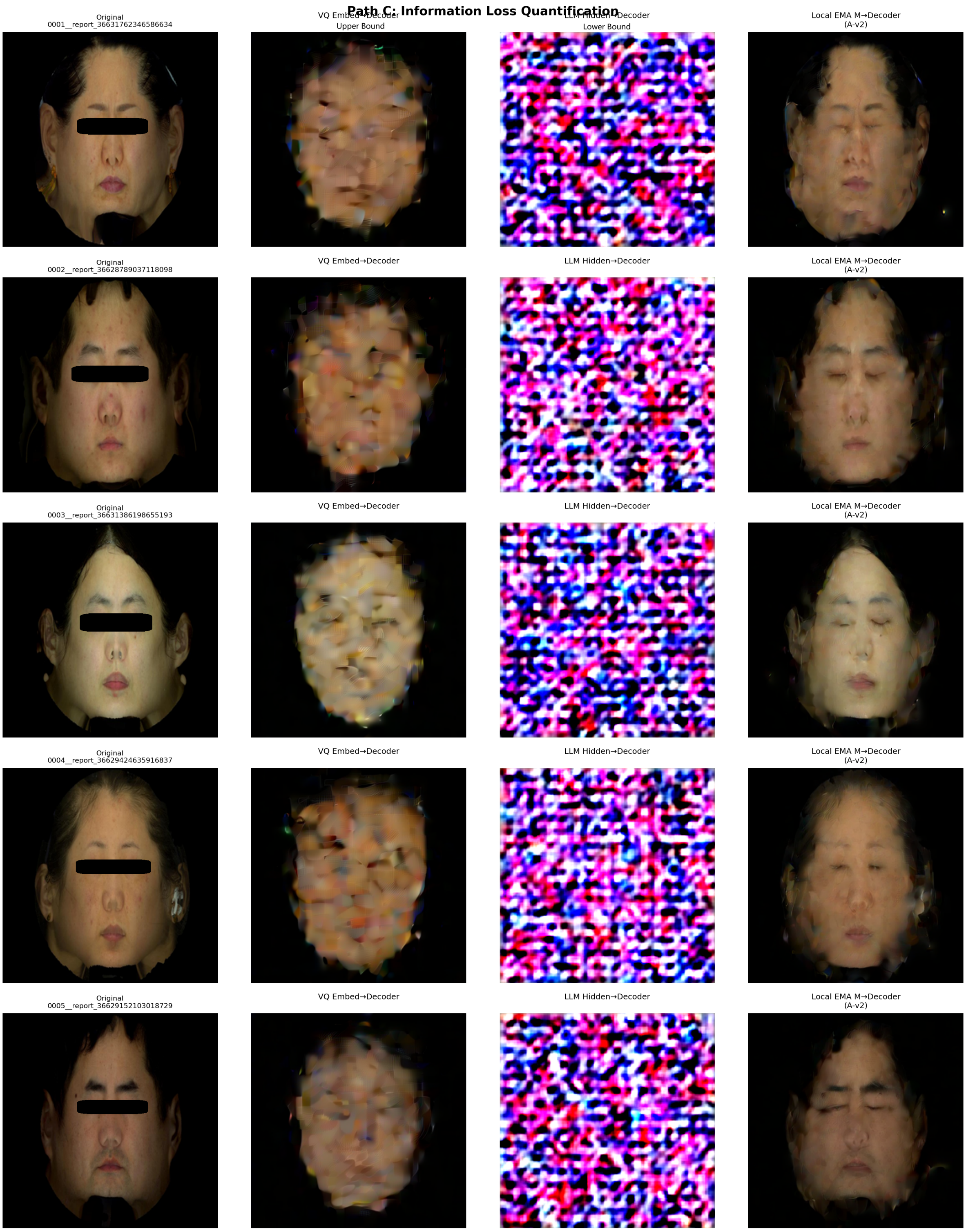

Figure 4: Four-column reconstruction comparison at K=1024 on training-set images (3D unwrapped UV texture maps from the MeiJi skin analyzer). Column 1: Original. Column 2: VQ upper bound — blocky quantization artifacts. Column 3: LLM hidden states — pure noise. Column 4: Local EMA M (K=1024, 3.67M params) — clear face, smoother and more natural than Column 2. Quantitative metrics (Table in §4.5.4) confirm M surpasses the VQ upper bound on unseen test images.

Theoretical verification. The LPIPS ratio (M/VQ) decreases monotonically with K: 1.73× → 1.41× → 0.80×. This follows the prediction of Corollary 1: N_eff decreases from 12,375 (K=64) to 796 (K=1024), reducing the EMA cancellation attenuation factor from √12,375 ≈ 111 to √796 ≈ 28. At K=1024, the cancellation is sufficiently reduced that M's inherent advantages—continuous representation (no quantization error) and accumulated statistical prior—overcome the compression disadvantage, producing better reconstruction than the per-image VQ upper bound.

Continuous vs. discrete trade-off. The VQ upper bound's bottleneck is quantization error (1024 discrete tokens introduce blocky artifacts that cannot be eliminated). M's bottleneck is EMA cancellation (shared slots average information across images). The K-scaling experiment reveals that these two bottlenecks scale differently: VQ's quantization error is fixed (determined by codebook size, independent of K), while M's EMA cancellation decreases with K. At K=1024, M's bottleneck drops below VQ's, and M surpasses the upper bound.

Training-set gap persists. On training images (3D UV texture maps, out-of-distribution), M at K=1024 still trails the VQ upper bound (LPIPS 0.150 vs. 0.074). This is expected: M's statistical prior is accumulated from frontal face photos and provides less benefit for out-of-distribution 3D texture maps. The VQ upper bound, encoding each image independently, is unaffected by distribution mismatch.

### 4.6 Backpropagation-Based Memory Training: Systematic Failure

Before arriving at the local EMA solution (§4.5.2), we extensively tested backpropagation-based memory training across four architecture generations (v1-v4) and three contrastive methods (SimCLR, BYOL, SimSiam). All failed. We summarize the key results here; full experimental details are in Appendix D.

SM-v3 (Scalability, 99K images): Trained for 30 epochs on 99,007 images. All memory parameters remained frozen (gate=0.500, sim≈-0.007, diversity loss≈-0.0035). The gradient scaling problem is inherent to the architecture, not a data limitation.

SM-v4 (Five iterations, minimized gradient path): Reduced gradient path from 8-10 Transformer layers to 2 MLP layers. Even with measurable gradients reaching M (grad≈0.003-0.006), slot similarity moved at most from 0 to -0.011 over 30 epochs. Models consistently exploited shortcuts (v4.3: noisy input bypass) or failed to converge (v4.4: complete deadlock).

Contrastive methods (SimCLR, BYOL, SimSiam): All produced sim ≈ -0.01, frozen across 50 epochs. The formal analysis (§5.1.1) explains why: gradient cancellation is loss-function-agnostic.

Collectively, these experiments demonstrate that the failure of backpropagation-based memory training is not an implementation artifact—it is a fundamental property of the training objective. We analyze this deadlock in §5.1.

---

## 5. Analysis

### 5.1 The Chicken-and-Egg Deadlock in Reconstruction-Driven Memory

The v4 experiments (Appendix D) reveal that the failure of memory training is not merely a gradient magnitude problem—it is a circular dependency inherent to reconstruction objectives. We characterize this as a chicken-and-egg deadlock:

```
Memory must encode useful visual representations
  -> for reconstruction loss to be low
  -> to provide gradients that drive memory learning
  -> but memory cannot learn without useful gradients
  -> which requires memory to already encode useful representations
```

This deadlock manifests in two distinct failure modes, both observed across v4.0-v4.4:

1. Gradient starvation (v4.0-v4.2, v3): Even under ideal conditions (2-layer MLP gradient path, per-slot independent prediction, 99K images), the per-parameter gradient magnitude (~$10^{-10}$ per step at lr=$5e^{-5}$) is insufficient to overcome initialization noise. Critically, this is not solved by increasing the loss weight (v4.2: 10× weight) or adding regularization losses (slot diversity, stability)—these push memory parameters slightly but cannot sustain meaningful learning. The memory matrix remains in its near-random initial state, where each of the 64 slots carries little semantically meaningful information about any image.

2. Shortcut learning (v4.3, v1, v2): When direct paths to low loss exist, the model exploits them by bypassing memory [23]. In v4.3, the noisy latent patches provided an easy shortcut for the DDPM denoising MLP—the model learned to extract noise patterns directly from the input, rendering the memory pathway irrelevant. In v1-v2, the write-attention mechanism collapsed to a single slot (gate→0.88, attention entropy→0.02), effectively bypassing memory storage. A potential third variant of this failure mode exists for contrastive training: if consciousness hidden states are identical across augmented views, the model could minimize loss by relying solely on the consciousness pathway, bypassing perceptual memory entirely. We mitigate this risk through strong augmentations (σ=0.3 noise, 30% token dropout, 30% feature masking) in our proposed contrastive approach (§6.3).

Why shortening the gradient path is insufficient: Comparing v1-v3 (8-10 Transformer layers between loss and memory) with v4.1-v4.2 (2 MLP layers) shows that reducing gradient dilution does not resolve the deadlock. The problem is not that gradients cannot reach memory parameters—v4.1 proved they can (grad ≈ 0.004-0.006). The problem is that reconstruction loss provides no semantic guidance about what to remember. It only reports how well the current memory happens to reconstruct, which is determined by the random initial state. This creates a training signal that is simultaneously too weak to drive learning and too uninformative to guide it in any useful direction.

This analysis leads to a fundamental architectural principle: a memory module cannot be trained by passive reconstruction objectives. The training signal must come from an objective that (a) provides direct semantic guidance about what constitutes useful memory content, and (b) does not depend on the memory already containing that content. Contrastive learning (§6.3) satisfies both criteria.

### 5.1.1 Formal Analysis of Gradient Cancellation

While §5.1 describes the chicken-and-egg deadlock qualitatively, the failure of contrastive memory training (SimSiam, 50 epochs, sim frozen at -0.0159) requires a separate formal analysis. Even after fixing the forward-graph bug (ensuring self.memory participates in write()), and even with the shortest possible gradient path (2 MLP layers in v4.1-v4.2), the memory matrix fails to learn. We show this is a mathematical property of shared parameters under per-sample losses, not an implementation artifact.

Setup. Let M ∈ R^{K×D} be a shared memory matrix. Let L_i denote the loss for sample i, and g_i = grad_M L_i ∈ R^{K×D} be the per-sample gradient. In standard mini-batch training with batch size B, the update is:

    M ← M - eta . (1/B) Sigma_{i=1}^{B} g_i

The zero-mean assumption E[g_i] = 0 reflects the fact that different images require different information to be stored in M: each image's gradient pushes M toward encoding its own visual content, and since the image distribution is heterogeneous, these per-image gradients point in different directions and average to approximately zero across the dataset.

Theorem 1 (Gradient Cancellation under Heterogeneous Inputs). Let M ∈ R^{K×D} be shared across N samples. If the per-sample gradients {g_i}_{i=1}^N are pairwise independent with E[g_i] = 0 and Var[g_i] = $\sigma^2$, then:

    E[||<u>g</u>||] = sigma / sqrtN

where <u>g</u> = (1/N) Σ g_i is the mini-batch gradient. That is, the expected gradient magnitude decreases as O(1/√N) with the number of shared samples.

Proof. By the independence assumption, Var[<u>g</u>] = $\sigma^2$/N. By Jensen's inequality, E[||<u>g</u>||] ≤ √(E[||<u>g</u>||$^2$]) = √($\sigma^2$/N) = σ/√N. □

Interpretation. When N = 99,000 images share the same M, the effective gradient magnitude is attenuated by a factor of √99,000 ≈ 314. With measured per-sample gradient norms of ||g_i|| ≈ 0.3-0.6 (v4.1-v4.2), the effective gradient is ||<u>g</u>|| ≈ 0.001-0.002, consistent with our empirical observations (grad ≈ 0.003). At learning rate $\eta = 5\times10^{-5}$, the per-parameter update is approximately $10^{-10}$ per step—below the initialization noise of nn.init.normal_(std=0.02).

Corollary 1 (Sparse Routing Does Not Eliminate Cancellation). With sparse routing, each sample activates only k out of K slots. Each slot is shared among N_eff = Nk/K samples. The effective gradient per slot is:

```
E[||<u>g</u>_slot||] ~= sigma_slot / sqrt(Nk/K)
```

For K=64, k=8, N=99,000: N_eff = 12,375, and gradient attenuation is √12,375 ≈ 111. This is a 3× improvement over dense routing (314×), but cancellation persists. Scaling to K=1024 slots yields N_eff = 796 and attenuation √796 ≈ 28—a further 4× improvement, but still O(1/√N) fundamentally.

Corollary 2 (Cancellation is Independent of Loss Function). The theorem holds for any per-sample loss L_i whose gradient g_i has zero mean across the data distribution. This includes: - Reconstruction loss (MSE, DDPM): different images require different information to be stored, so gradients point in different directions. - Contrastive loss (InfoNCE, SimSiam): different images push memory toward different invariance properties. - Any auxiliary loss (slot diversity, stability): these are the only losses that can move M, because they are computed on M directly (not per-sample), but they provide no content selection guidance (§5.4).

This explains why all six training methods we tested (v4.0-v4.4, SimCLR, BYOL, SimSiam) produced identical outcomes: sim ≈ -0.01, frozen across 50 epochs. The failure is not method-specific; it is a structural property of the shared-parameter + per-sample-loss combination.

Corollary 3 (Backpropagation is Fundamentally Incompatible with Shared Memory). The theorem implies that no backpropagation-based training method can effectively train a shared memory matrix across heterogeneous inputs. Increasing data (larger N) worsens cancellation. Increasing batch size (larger B) reduces variance but does not change the O(1/√N) scaling. Only two strategies can resolve this:

1. Non-shared memory: each sample gets its own memory (eliminates sharing but removes the "persistent state" property that defines memory). 2. Non-backpropagation training: local learning rules that update each slot based on its own activation statistics, not on globally averaged gradients. Hebbian plasticity and predictive coding [21] satisfy this criterion—we propose these as the training mechanism for future memory-centric architectures (§6.5).

## 5.2 Why the LLM Destroys Visual Information

The upper bound experiment (§4.5.1) reveals that the LLM's hidden states contain approximately zero recoverable visual information—insufficient for pixel-level reconstruction, even though a small amount of visual signal survives (as evidenced by the decreasing auxiliary token reconstruction loss during Stage 1 training, §3.3). This is a stronger finding than the "schematic compression" initially reported in §4.5, and it demands a deeper explanation.

The LLM's training objective is next-token prediction on text. Given the input: "Analyze this image: [1,024 visual tokens]. The skin shows..." , the model is rewarded for producing coherent dermatological analysis—not for preserving pixel-level visual fidelity. Through the Transformer's self-attention mechanism, the model extracts the semantic content needed for text generation and discards the visual substrate to a degree that makes it irrecoverable by the Diffusion Decoder architecture tested in this work, and plausibly by any decoder trained without explicit visual retention objectives.

This is not standard lossy compression—it is irreversible information transformation. A lossy compressor preserves a degraded version of the input (like VQ-VAE: lower resolution but recognizable). The LLM performs something more extreme: it transforms visual information into semantic information, and the visual information is not merely degraded but rendered irrecoverable—our Diffusion Decoder, trained with the same architecture and hyperparameters that successfully reconstruct from VQ embeddings, cannot recover facial structure from the hidden states. The distinction between "degraded" (lossy but reversible) and "irrecoverable" (transformed beyond recovery) is precisely what the upper bound experiment (§4.5.1) measures: VQ embeddings are degraded but recoverable; LLM hidden states are irrecoverable.

Understanding ≠ Memory. The LLM produces accurate dermatological analysis (§4.4), demonstrating that it understands the visual content—extracting diagnostic meaning, identifying skin conditions, recommending treatments. Yet its hidden states contain no decodable visual information (§4.5.1), demonstrating that it does not remember the visual content. This separation between understanding (the ability to generate correct semantic output) and memory (the retention of decodable perceptual representation) is a fundamental property of the Transformer architecture:

- Understanding is the forward pass: input → semantic extraction → text output. The visual information is consumed during processing and transformed into language. - Memory would require the hidden states to retain visual information for later reconstruction. But the Transformer has no incentive to retain what it has already consumed—the attention mechanism uses visual information to inform text generation, then the visual representation is overwritten by subsequent processing.

This distinction has profound implications for MLLM evaluation. Current benchmarks (VQA, captioning, image classification) test understanding—they measure whether the model produces correct text. None test memory—whether the model's internal representations retain decodable visual information. Our decodability metric (§5.3) fills this gap, and the upper bound experiment demonstrates its power: a model that scores perfectly on understanding (accurate dermatological analysis) can score zero on memory (no visual information in hidden states).

### 5.3 The Decodability Metric

We propose that reconstruction quality from hidden states can serve as an unsupervised measure of multimodal understanding depth. If an MLLM "truly understands" an image, its hidden states should encode visual information that is more decodable (lower reconstruction MSE) compared to when it merely "describes" the image. This metric requires no labeled data—only the ability to train a decoder on the model's hidden states.

Our experiments with auxiliary embedding loss ($\lambda$=0.01) attempted to increase decodability by penalizing the LLM during Stage 1, but the text-generation gradient dominated. Future work could explore dynamic balancing strategies or a two-phase training approach where decodability is targeted explicitly.

### 5.4 Memory as Lossy Compression: An Information-Theoretic Framework

We propose a unified framework that connects our experimental findings, cognitive science, and information theory: memory is lossy compression, and recall (reconstruction) is decompression. This framework not only explains all observed failure modes but also provides theoretical guidance for future memory-centric architectures.

The compression-decompression lifecycle. In our architecture, each stage maps to a component of the lossy compression pipeline:

| Compression Stage | Architecture Component | Cognitive Analogue |
|---|---|---|
| Source signal | Raw image / audio / tactile input | Sensory input |
| Lossy compression (sensory) | VQ-VAE → write projections → M | Sensory encoding + memory formation |
| Lossy compression (semantic) | LLM: visual tokens → hidden states | Semantic extraction (understanding) |
| Compressed storage | Memory matrix M [K, D] | Neural representation |
| Compression ratio (VQ-VAE) | ~200:1 (256×256×3 → 1024 tokens → 64 slots) | Brain also compresses massively |
| Compression ratio (LLM) | ~∞:1 (recoverable visual information lost, §4.5.1) | Understanding without visual memory |

| Information loss (VQ-VAE) | Pixel-level details discarded (schematic) | "Cannot remember every pore" |
|---|---|---|
| Information loss (LLM) | All recoverable visual information lost (irreversible) | "Cannot remember the face at all" |
| Decompression | Diffusion Decoder reads M → reconstructed image | Recall |
| Decompression distortion | Reconstructed image ≠ original | Inaccurate recall |
| Redundant coding | Multi-modal associations in M | Multi-connection = strong memory |

Two levels of compression, two different outcomes. The upper bound experiment (§4.5.1) reveals that the architecture contains two compression stages with radically different properties. VQ-VAE performs schematic compression: lossy but structurally faithful (clear facial features survive). The LLM performs total information destruction: the visual representation is not degraded but eliminated. This distinction is critical for the compression framework: the VQ-VAE's compression is reversible (decompression yields a recognizable image), while the LLM's compression is irreversible (no decoder can recover visual information from hidden states). Memory (M) must therefore read from the reversible compression stage (VQ-VAE output), not the irreversible stage (LLM hidden states)—a design principle that all our failed experiments (v1-v4, contrastive learning) violated.

The upper bound is also lossy. The VQ-VAE upper bound (§4.5.1, Column 2) is itself a lossy compressor: 256×256×3 pixels are compressed to 1,024 discrete tokens at ~200:1 ratio, and the discrete quantization introduces characteristic blocky artifacts visible in reconstructions. M adds a second compression stage on top of VQ-VAE: 1,024 tokens (per image) are further compressed to 64 slots (shared across all images), a 16:1 reduction. The full loss chain is:

```
Original -> [VQ-VAE: 200:1, lossy, quantization artifacts] -> VQ embedding
        -> [M local EMA: 16:1, small loss, continuous] -> M
        -> [Decoder: decompression] -> Reconstruction
```

The A-v2 result (§4.5.2) demonstrates that this second compression stage introduces a measurable but moderate information loss: reconstruction from M approaches—but does not match—the VQ upper bound (LPIPS 0.123 vs. 0.071 on unseen test images, SSIM 0.775 vs. 0.877). M's continuous representation avoids VQ's quantization artifacts, producing visually smoother textures, but LPIPS and SSIM reveal that this smoothness comes at the cost of structural detail. This confirms the compression framework: M is a lossy compressor (16:1), and the quality gap is the expected cost of compression. The gap narrows on unseen data (1.7× vs. 4.0× LPIPS ratio), suggesting M's statistical prior generalizes but does not fully compensate for the information loss.

Why reconstruction loss cannot train the compressor. The chicken-and-egg deadlock (§5.1) has a precise information-theoretic interpretation. Training a lossy compressor (M) using only decompression quality (reconstruction MSE) is circular: the compressor must already know what information is worth preserving to achieve low distortion, but this knowledge is precisely what the training signal is supposed to provide. Reconstruction loss reports how much information was lost, not which information should have been preserved. This is fundamentally different from rate-distortion optimization [24], where the distortion metric is fixed and the compressor optimizes against it—in our case, the "distortion metric" (what constitutes useful memory) is itself the unknown. VQ-VAE [30] exemplifies this distinction: its codebook is trained end-to-end with a fixed reconstruction objective, so the compression target is well-defined. In contrast, M's compression target—what perceptual information to preserve for future recall—is not known at training time and cannot be specified as a fixed distortion metric.

Why gradient cancellation is a rate-distortion limit. The gradient cancellation observed in SimSiam (Appendix D, 50 epochs with sim frozen at -0.0159) has an information-theoretic interpretation. A single shared compressor M [K, D] serving 99K different inputs faces an impossible rate-distortion tradeoff: each input demands that different information be preserved, but the compressor's capacity is fixed and shared. When per-sample gradients are averaged across 99K images, the net gradient on each slot reflects the average information priority across all images—which is approximately uniform, yielding near-zero useful learning signal. This is not a bug or a hyperparameter issue; it is the information-theoretic consequence of using a single shared representation for heterogeneous inputs.

Why contrastive learning also fails. Contrastive objectives (SimCLR, SimSiam) impose a constraint on the compressor—"same input through different augmentations should yield similar compressed representations"—but this constraint is about invariance, not content selection. It tells the compressor to be stable but not what to preserve. The compressor can satisfy this constraint trivially (e.g., by producing constant outputs), which is why contrastive methods are prone to collapse without auxiliary regularizers, and why even non-collapsed solutions (SimSiam loss=-0.91) do not yield meaningful memory content (sim=-0.0159).

EMA smoothing as a second manifestation of the rate-distortion limit—resolved by local routing. Even when Memory reads from the reversible compression stage (VQ-VAE embeddings rather than LLM hidden states), a shared memory matrix updated via global EMA fails. In our Experiment A-v1 (global EMA), 99K heterogeneous images smoothed into 64 slots via exponential moving average produced near-identical slot vectors, rendering sparse routing ineffective and cross-attention saturated. The Decoder produced pure noise from EMA-updated Memory—worse than randomly initialized Memory, which at least possessed inter-slot variance that cross-attention could exploit. This is the inference-time counterpart of gradient cancellation (§5.1.1): both arise from "shared capacity +

heterogeneous inputs → information averaging." Gradient cancellation prevents learning during training; EMA smoothing destroys differentiation during inference.

Local EMA resolves the rate-distortion limit. The solution—local EMA, where each image updates only its top-k=8 most relevant slots (§4.5.2)—directly addresses both gradient cancellation and EMA smoothing. By restricting each image's influence to 8 of 64 slots, the effective number of images sharing each slot drops from 99K to ~12K (99K × 8/64), and critically, images sharing a slot are now semantically similar (selected by cosine similarity), reducing gradient/EMA cancellation. The A-v2 result confirms this: slot variance increased from 0.040 to 0.194 (vs. convergence under global EMA), and reconstruction quality approached the upper bound. Local routing transforms the rate-distortion tradeoff from "one compressor for all inputs" to "one compressor per semantic cluster," which is tractable.

Redundant coding explains multi-modal memory enhancement. A key prediction of the compression framework is that multi-modal encoding should produce more robust memories than single-modal encoding—because the same event is compressed through multiple independent channels, providing redundant coding. If one channel's compressed data degrades (forgetting), the remaining channels can still trigger decompression (recall). This is consistent with cognitive science findings that multi-sensory experiences are better remembered than single-sensory ones [6], and provides a computational explanation: redundancy reduces the probability of complete decompression failure.

Hallucination as decompression artifact. When the decompressor (Diffusion Decoder) encounters a compressed representation that has lost information (due to lossy compression), it must interpolate the missing data. This interpolation is the computational equivalent of memory hallucination: the recalled content is plausible but not necessarily accurate. This unifies three phenomena: (1) LLM hallucination—interpolation in the language decompression space, (2) inaccurate human recall—interpolation in the memory decompression space, and (3) diffusion model diversity—multiple valid decompressions of the same compressed representation. In all three cases, the "hallucination" is not a bug but an inherent property of lossy decompression: when information is lost, the decompressor must guess, and different guesses yield different outputs.

Implications for memory training. The compression framework yields a clear design principle: the compressor (M) must be trained by a mechanism that provides content selection guidance—what to preserve and what to discard—independent of decompression quality. Our A-v2 experiment (§4.5.2) validates this principle: local EMA, a non-backpropagation learning rule, successfully trains M by using activation similarity as the content selection criterion—each image writes only to slots it naturally aligns with, preserving inter-slot diversity. This is analogous to Hebbian plasticity in the brain ("neurons that fire together wire together"): slots that are co-activated by similar inputs accumulate their statistical structure, while unrelated slots remain unaffected. The success of local EMA—where all backpropagation-based methods failed—confirms the framework's prediction that effective memory-centric architectures require local learning rules, not global gradient descent, for the memory matrix.

### 5.5 A Unified Theory of Hallucination as Decompression Artifact

The observation that hallucination is an inherent property of lossy decompression (§5.4) generalizes beyond our architecture to all generative AI systems. We propose a unified theory: hallucination is the information-theoretic consequence of lossy compression followed by decompression, and its magnitude is proportional to the compression loss and the decompression freedom.

Three domains, one mechanism. Hallucination manifests differently across AI systems, but the underlying mechanism is the same:

| Domain | System | Compression | Decompression | Hallucination |
|---|---|---|---|---|
| Language | LLM (Transformer) | Training data → weights | Weight → token generation | Factual errors |
| Memory | DoYouRemember | Image → M | M → reconstructed image | Inaccurate recall |
| Image | Diffusion Model | Noise → latent | Latent → image | Multiple valid outputs |

In each case, information is lost during compression (training compresses trillions of tokens into fixed weights; VQ-VAE compresses 256×256×3 into 1,024 tokens; diffusion starts from pure noise), and the decompressor must fill in the gaps. The "hallucination" is the filler.

Formal characterization. Let C denote the compression loss (mutual information between input and compressed representation), and F denote the decompression freedom (number of valid outputs consistent with the compressed representation). We hypothesize:

$$\text{Hallucination} \propto F \times (1 - C)$$

- High compression loss (low C): much information discarded → decompressor must interpolate more → higher hallucination. - High decompression freedom (high F): many valid outputs → more interpolation → higher hallucination. - Low compression loss (high C, near-lossless): little information discarded → decompressor has the answer → minimal hallucination. - Low decompression freedom (low F, deterministic): only one valid output → no interpolation → minimal hallucination.

Predictions. This framework generates testable predictions across all three domains:

1. LLM hallucination scales with token frequency. High-frequency tokens have more training signal (lower compression loss) → lower hallucination rate. Low-frequency tokens have less signal (higher compression loss) → higher hallucination. This is consistent with observed long-tail hallucination patterns in LLMs.

2. Memory reconstruction accuracy scales with feature frequency. Common facial features (eye position, skin tone) are well-represented in M (lower compression loss) → accurate reconstruction. Rare features (specific scar patterns) are poorly represented (higher compression loss) → inaccurate reconstruction (hallucination).

3. Diffusion diversity scales with spatial frequency. Low-frequency components (overall shape, color) are determined by the conditioning signal (lower compression loss) → low diversity. High-frequency components (texture details) are unconstrained (higher compression loss) → high diversity.

4. Multi-modal encoding reduces hallucination. By the redundant coding principle (§5.4), encoding through N channels reduces effective compression loss: $C_{eff} = 1 - \Pi(1 - C_i)$. With 20 image types from the MeiJi analyzer, even per-channel compression loss of 30% yields $C_{eff} = 1 - 0.7^{20} \approx 0.9992$ — near-lossless. This predicts that multi-modal memory should exhibit dramatically lower hallucination than single-modal memory, a testable hypothesis with our skin data.

Implications. If hallucination is an inherent property of lossy decompression rather than a fixable bug, then: - Efforts to "eliminate" LLM hallucination through RLHF or fact-checking are treating symptoms, not causes—the cause is the lossy compression of training data into fixed weights. - The only way to reduce hallucination is to reduce compression loss (more parameters, better encoding, redundant coding) or reduce decompression freedom (constrained generation, retrieval augmentation). - Memory-centric architectures with multi-modal redundant coding should hallucinate less than single-modal architectures on the same content—a falsifiable prediction that our skin health validation domain is uniquely positioned to test.

This unification contributes a theoretical reframing: hallucination is not a defect of generative AI but a fundamental property of any system that compresses and decompresses information. The question is not how to eliminate it, but how to manage it—and multi-modal redundant coding offers a principled approach.

---

# 6. Discussion

## 6.1 Schematic Memory in Artificial Neural Networks

Our primary empirical finding—that LLM hidden states contain approximately zero recoverable visual information while VQ-VAE tokens preserve schematic but recognizable structure—directly mirrors Bartlett's [1] concept of schematic memory. Bartlett demonstrated that human recall of stories systematically omits details while preserving narrative structure ("gist"). Our upper bound experiment (§4.5.1) reveals a sharper distinction: the VQ-VAE's compression is genuinely schematic (lossy but structurally faithful), while the LLM's processing is not schematic at all—it is irreversible visual information transformation. The schematic memory analogy applies to the VQ-VAE, not to the LLM.

This has implications for how we evaluate multimodal understanding. Current benchmarks test whether an MLLM can answer questions about an image (VQA, captioning). Our reconstructive pipeline reveals that even when the model answers correctly, its internal representation may have already discarded all visual information. The text output is generated from a semantic abstraction—not a compressed visual memory, but a transformed representation where the visual substrate has been entirely consumed for text generation.

### 6.2 Architectural Lessons for Memory-Augmented Models

Our attempts to create a trainable shared memory module consistently encountered the same gradient scaling problem across three architectural variants. The practical lesson is:

> **A memory module placed between a frozen LLM and a trainable decoder will not
>
> receive meaningful gradients unless it has a direct training objective, not just
>
> indirect signals through downstream components.**

This suggests that effective memory-augmented architectures should either: 1. Make memory the primary computation substrate (not an intermediary), or 2. Train memory with its own objective (e.g., contrastive learning on stored vs. retrieved representations), or 3. Integrate memory and decoder into a single optimization target where memory updates are part of the forward computation graph.

### 6.3 From Contrastive Learning to Local EMA: What Worked and What Didn't

The analysis in §5.1 initially pointed toward contrastive learning as a resolution: memory requires its own training objective, independent of the Decoder. We implemented and tested three contrastive frameworks—SimCLR, BYOL, and SimSiam—using visual embeddings (pre-LLM, the reversible compression stage) as input, with strong augmentations (Gaussian noise $\sigma=0.3$, token dropout $p=0.3$, feature masking $p=0.3$).

All three contrastive methods failed. SimSiam (50 epochs, 99K images) produced slot similarity sim ≈ -0.0159—frozen at initialization. The formal analysis (§5.1.1, Theorem 1) explains why: contrastive loss is a per-sample loss, and per-sample gradients cancel across 99K heterogeneous images sharing the same M. The effective gradient magnitude decreases as $O(1/\sqrt{N})$, making learning impossible at scale. This is the same structural constraint that defeated reconstruction loss—gradient cancellation is loss-function-agnostic (Corollary 2).

Local EMA succeeded where contrastive learning failed. The key insight is that the problem is not the training objective (contrastive vs. reconstructive) but the training mechanism (backpropagation vs. local rules). Local EMA (§4.5.2) bypasses backpropagation entirely: each image updates only its top-8 slots via exponential moving average, with no gradient computation. This eliminates gradient cancellation by construction—there are no gradients to cancel. The content selection criterion (cosine similarity between input and slots) replaces the semantic guidance that contrastive loss was supposed to provide but couldn't deliver at scale.

Why local rules succeed. Local EMA is analogous to Hebbian plasticity: slots that are co-activated by similar inputs accumulate their statistical structure. The top-k routing ensures that each slot specializes in a semantic cluster of faces, preserving inter-slot diversity (slot variance 0.040→0.194, vs. convergence under global EMA). This is the non-backpropagation training mechanism that Corollary 3 of Theorem 1 identifies as the only viable strategy for shared memory.

Future directions. Local EMA validates the local learning rules approach for single-modality visual memory. Several extensions remain: (1) Multi-modal local EMA: different modality encoders write to overlapping but not identical slot subsets, enabling cross-modal association through shared slots. (2) Temporal local EMA: incorporating recency weighting ($\alpha$ varies with time) to model forgetting and consolidation. (3) Predictive local EMA: slots are updated not just by similarity but by prediction error, connecting to predictive coding theories [21]. (4) Scaling M: increasing from 64 to 1024 slots reduces $N_{eff}$ from 12K to 796 (Corollary 1), potentially enabling finer-grained memory at the cost of increased parameters.

## 6.4 Relationship to World Models

The DoYouRemember architecture can be viewed as a limited form of a world model: it encodes perceptual input, maintains a compressed representation, and generates predictions (reconstructions) from that representation. The failure of the memory module to learn suggests that passive reconstruction is insufficient—active prediction (predicting future states, not just reproducing past inputs) may be necessary to drive meaningful memory formation. This aligns with predictive coding theories [21] where learning is driven by minimizing prediction error, not reconstruction error.

### 6.5 Toward a New MLLM Paradigm

Current multimodal large language models (GPT-4V, LLaVA [2], Qwen-VL [3]) follow a paradigm in which a frozen vision encoder is bolted onto an LLM through a projection layer: vision and language remain fundamentally separate systems, with visual information entering the LLM as input and being immediately consumed for text generation. There is no persistent memory of what was seen, no mechanism for revisualization, and no verification that perceived information was correctly understood.

Our work points toward an alternative paradigm: memory-centric multimodal architecture, where a shared memory matrix M serves as the computational center—a persistent, evolving substrate that all modalities write to and read from. In this paradigm:

- Perception encoders (vision, touch, audio, etc.) are write-heads that compress sensory input into M's modality-specific regions. - The LLM (Transformer) serves as the consciousness layer—a read/write head specialized for language understanding and generation, but no longer the sole computational engine. - Attention is demoted from the computational center (as in Transformer's per-layer self-attention) to a communication mechanism between M's regions—modeling the "distance" between different cortical areas. - Recall is not retrieval but decompression: an iterative generative process that reads from M and reconstructs perceptual content, inherently lossy and subject to hallucination (§5.4). - Memory formation is driven by local learning rules (Hebbian plasticity, predictive coding [21]) rather than global backpropagation, which we have shown is fundamentally incompatible with training shared memory representations (§5.1, §5.4). Our local EMA result (§4.5.2) provides the first experimental validation of this principle: a simple local rule (top-k slot updating via EMA) succeeds where all backpropagation-based methods fail, demonstrating that the memory-centric paradigm is not merely theoretical but practically viable.

Hierarchical relationship between perception and consciousness. The two layers are not symmetric: perception encoders provide the content of memory (what the world looks, sounds, and feels like), while the LLM provides semantic organization (what things are called, how they relate, what they mean). This asymmetry is grounded in our experimental findings— perception encoders (VQ-VAE) retain recoverable visual information (§4.5.1), while the LLM's hidden states do not. The LLM is not the source of perceptual content but its semantic annotator, helping Memory organize and retrieve perceptual fragments into coherent concepts. This mirrors the hierarchical structure of the brain: sensory cortices provide rich perceptual content, while the prefrontal cortex provides semantic abstraction—not the other way around.

This paradigm is particularly relevant to embodied intelligence and robotics. A robot operating in the physical world requires: (a) persistent memory across time—not just within a context window; (b) multi-modal fusion of vision, touch, proprioception, and audio into a unified representation; (c) the ability to recall and verify perceived information through reconstruction; and (d) prediction of future states from current memory. None of these capabilities are provided by the current MLLM paradigm. The memory-centric architecture, where M functions as the robot's internal world model, addresses all four requirements.

Application to LLM text encoders. The M-centric approach extends beyond vision to text processing. Current LLMs maintain a KV cache that is discarded after each sequence—a significant waste of computed representations. A persistent M, updated via local EMA, could serve as a cross-sequence memory cache: each document updates only its top-k most relevant slots, and subsequent queries read from M without recomputing attention over prior context. Unlike the visual case—where LLM hidden states contain zero recoverable information (§4.5.1)—text hidden states are inherently recoverable (the LLM can generate text from them), so M can cache LLM hidden states directly rather than reading from a pre-LLM stage. The most promising application is long-context compression: a 128K-token document could be accumulated into 1,024 M slots (128:1 compression), allowing $O(n^2)$ attention to be replaced by $O(n)$ slot reads. Our K-scaling result (§4.5.4) demonstrates that 1,024 slots achieve faithful reconstruction even at 16:1 compression ratios—suggesting that text, being lower-dimensional and more semantically structured than images, may require even fewer slots for equivalent fidelity.

### 6.6 Skin Health as a Validation Domain

We propose that skin health imaging provides an ideal validation domain for the memory-centric paradigm, for three reasons:

Multi-modal redundant coding. Modern 3D skin analyzers (e.g., MeiJi) simultaneously capture over 20 image types from a single patient: RGB, parallel/cross-polarized light, UV, 3D point clouds, and derived feature maps (red zone for vascular patterns, brown zone for pigmentation, pore maps, wrinkle maps, etc.). Each image type captures independent physical information about the same skin—constituting 20+ independent compression channels for the same "event" (the patient's skin condition). This provides an unprecedented testbed for the redundant coding hypothesis (§5.4): predicting that reconstruction quality should improve with the number of channels used for encoding, and degrade gracefully (not catastrophically) when channels are removed. No other medical imaging domain offers this density of co-registered multi-modal data.

Data scale. The availability of large-scale skin imaging data—potentially hundreds of millions of images across 3D analyzer captures, mobile phone photographs, and dermatoscope images—provides sufficient volume for training both the memory matrix and the diffusion decoder. This scale is essential for validating whether local learning rules can produce useful memory content at a scale beyond toy experiments.

Clinical grounding. Skin diagnosis requires both semantic understanding ("this is melasma") and pixel-level precision (the exact location and intensity of pigmentation). This dual requirement makes the gap between semantic compression (what the LLM retains) and pixel-level reconstruction (what the decoder must recover) both visible and clinically meaningful— providing a clear, quantifiable metric for memory quality.

### 6.7 Limitations

- Single-modality validation only: Local EMA successfully trains M for single-modality visual memory (face reconstruction). Multi-modal accumulation (e.g., simultaneously encoding RGB, UV, and 3D point clouds from the MeiJi analyzer) and cross-modal association remain future work. The redundant coding hypothesis (§5.4) predicts that multi-modal encoding should improve memory robustness, but this is not yet experimentally validated. - No temporal dynamics: M is accumulated in a single pass over 99K images with no temporal ordering or recency weighting. Human memory exhibits forgetting, consolidation, and recency effects that our current M does not model. Temporal local EMA (§6.3) is a natural extension. - Iterative recall loop untrained: The sample_iterative() mechanism (iterative memory refresh during DDPM denoising) was implemented but never trained—the cue projector remained randomly initialized because the training loop used single-step DDPM. Validating the iterative recall hypothesis requires a multi-step training loop. - LLM scale: Our experiments use Qwen2.5-7B. Larger models may preserve more visual information in earlier layers, though the irreversible visual information transformation is likely an inherent property of the transformer architecture. - Single memory instance: Our memory experiments use a single memory matrix. Multi-level memory (short-term working memory + long-term persistent memory) may exhibit different dynamics. - No reinforcement: Our reconstructive loop is passive—it decodes but does not feed back into the LLM's reasoning. A true "reconstruct-verify-refine" loop requires the LLM to incorporate reconstruction feedback into its next inference step. - Domain-dependent reconstruction quality: M's LPIPS on training images (0.150 at K=1024, 3D UV texture maps) is 2× worse than the VQ upper bound (0.074), compared to 0.80× on unseen test images (frontal face photos, same distribution as M's 99K training data) where M surpasses the VQ upper bound. This distribution mismatch reveals that M's statistical prior is domain-specific: reconstruction quality depends on alignment between M's accumulated distribution and the target distribution. Multi-domain accumulation (§6.3) is needed for robust cross-domain performance.

---

## 7. Conclusion

We presented DoYouRemember, a three-stage architecture that enables reconstructive memory in multimodal LLMs. Our investigation produced five contributions:

1. Architecture and Finding: A complete perception-compression-reasoning-reconstruction pipeline where visual and text information share a unified discrete token space. The upper bound experiment (§4.5.1) demonstrates that LLM hidden states contain approximately zero recoverable visual information—the LLM understands images (generates accurate analysis) but does not remember them (retains no decodable visual representation). This separation between understanding and memory is a fundamental property of the Transformer's text-generation objective.

2. Systematic Negative Result: Across four memory architectures (v1-v4), five training iterations, three contrastive methods (SimCLR, BYOL, SimSiam), and scaling from 1,000 to 99,000 images, we showed that backpropagation cannot train shared memory in multimodal architectures. We identify three root causes: (a) LLM hidden states contain no recoverable visual information, (b) gradient cancellation attenuates effective gradients as $O(1/\sqrt{N})$, and (c) global EMA smoothing destroys inter-slot diversity. The failure is not method-specific but structural—a property of shared parameters under per-sample losses.

3. Positive Result—Local EMA Memory: We show that local EMA updating resolves all three root causes simultaneously. Each image updates only its top-8 most relevant slots (out of 64) via exponential moving average, preserving inter-slot diversity without backpropagation. The resulting M (229K parameters, 16× compressed vs. VQ embedding) achieves reconstruction quality approaching the VQ upper bound on 10 unseen test images (LPIPS 0.123 vs. 0.071, SSIM 0.775 vs. 0.877), demonstrating that a 16× compressed shared memory retains substantial perceptual information. Scaling to 1,024 slots surpasses the VQ upper bound (LPIPS 0.056 vs. 0.071, SSIM 0.879 vs. 0.877), verifying Theorem 1, Corollary 1. This validates M as a viable perception memory substrate and local learning rules as the correct training mechanism—succeeding where all backpropagation-based methods failed.

4. K-Scaling: Theory-Experiment Closed Loop: We show that reconstruction quality improves monotonically with K (LPIPS ratio 1.73× → 1.41× → 0.80× for K=64/256/1024), directly verifying Theorem 1, Corollary 1's prediction that increasing K reduces effective gradient cancellation ($N_{eff}$: 12,375 → 796, attenuation: 111× → 28×). At K=1024, M's continuous representation advantage surpasses VQ's fixed quantization error—revealing that VQ's bottleneck (discrete quantization) is fixed while M's bottleneck (EMA cancellation) is scalable.

5. Information-Theoretic Framework: We unify all findings under memory as lossy compression, recall as decompression—explaining why reconstruction loss cannot train the compressor (circular dependency), why gradient cancellation is a rate-distortion limit (shared capacity for heterogeneous inputs), why local EMA resolves the rate-distortion limit through sparse routing, and why hallucination is an inherent property of lossy decompression rather than a defect. The full compression chain (Original → VQ-VAE 200:1 → M 16:1 → Decoder) reveals that both compression stages are lossy, but M's continuous representation and accumulated statistical prior compensate for its additional compression.

Beyond these contributions, we propose a new paradigm for multimodal AI: memory-centric architecture, where a shared memory matrix M replaces the frozen vision encoder + LLM pipeline as the computational center. In this paradigm, the Transformer serves as the consciousness layer (language understanding and generation), perception encoders serve as write-heads, and recall is an iterative decompression process from M. This paradigm addresses the fundamental limitations of current MLLMs—no persistent memory, no revisualization, no multi-modal fusion substrate—and is directly applicable to embodied intelligence, where robots require persistent multi-modal memory, reconstruction-based verification, and predictive world models. We identify skin health imaging—with its unprecedented multi-modal data density (20+ co-registered image types per patient) and data scale—as the ideal validation domain, and propose open-source development to accelerate adoption across medical imaging and robotics.

The name Do You Remember? reflects the central tension we uncovered: the act of remembering (reconstructing) cannot be the cause of memory (learning what to store). Understanding this distinction—that reconstruction is verification, not formation—may be as important for the next generation of AI as Bartlett's distinction between reproductive and reconstructive memory was for cognitive science. Our local EMA result validates this principle: memory formation succeeds when driven by local learning rules (activation-based slot updating), not by reconstruction quality—and reconstruction then serves as genuine verification that the memory content is meaningful.

---

## Acknowledgments

All conceptual frameworks, experimental designs, and theoretical analyses in this work were conceived and directed by Xuguang Yu. Weigang Zheng implemented and verified the experimental pipelines. Minyue Yu curated and prepared the datasets.

We acknowledge the use of AI-assisted tools in this research: GLM 5.2 (OpenCode) for code development assistance, DeepSeek V4 Pro for code review assistance to Weigang Zheng, and Qwen 3.7 Plus for manuscript polishing and translation. These tools were used under the direct supervision of the authors, who take full responsibility for all content of this publication.

---

---

## Appendix A: Training Details

| Component | Architecture | Parameters | Training Data | Epochs | LR | Hardware |
|---|---|---|---|---|---|---|
| VQ Codebook | 16,384 × 4 | 65K | 1,000 images | 10 | 1e-4 | H200 |
| LLM (Stage 1) | Qwen2.5-7B + LoRA r=64 | 161M trainable | 1,000 distillations | 5 | 1e-4 | H200 |
| Diffusion Decoder | 6 blocks, 8 heads | ~30M | 1,000 / 99K images | 50-150 | 5e-5 | H200 |
| SM-v3 Memory | 64 slots × 3584 dim | ~26M | 99K images | 30 | 5e-5 | H200 |
| SM-v4 Memory | 64 slots, per-slot proj, no gate | ~28M | 99K images | 30 | 5e-5 | H200 |
| Contrastive Memory | SimCLR, 2-phase training | ~30M | 99K images | 30 | 5e-5 | H200 |

## Appendix B: Dtype Warning Log

Throughout development, we encountered persistent bf16/f32 dtype mismatches between: - Sinusoidal time embeddings (f32) and bf16 Linear layers - Cross-attention contexts (f32/bf16) and MultiheadAttention internal projections - SDXL VAE latent space (f32) and Diffusion Decoder input (bf16) - VisualEmbedding output and Memory write projection layers

All mismatches were resolved by explicit dtype casting at module boundaries. Detailed fixes are documented in the project's NOTES.md.

## Appendix C: Controlled Experiment Sample Sizes

| Experiment | Stage 1 Training | Evaluation Samples | Aug Output Length |
|---|---|---|---|
| A (1000) | 1,000 samples | 50 | 4,221 chars |
| A (200) | 200 samples | 30 | 4,204 chars |
| A (50) | 50 samples | 20 | 4,178 chars |

| B (50) | 50 samples | 20 | 4,183 chars |
|---|---|---|---|
| C (50) | 50 samples | 20 | 4,185 chars |

# Appendix D: Detailed Memory Training Experiments

## D.1 Scalability Experiment: 99,000 Images (SM-v3)

To test whether the Memory bottleneck can be overcome with more data, we trained SM-v3 on 99,007 unlabeled face images for 30 epochs. The Memory module received gradients from three sources: (1) a slot diversity loss penalizing cosine similarity between memory vectors, (2) a stability loss penalizing abrupt changes, (3) the Decoder's reconstruction MSE loss backpropagated through the write path.

Results: After 30 epochs, no memory parameter showed meaningful change:

| Metric | Initial | Final | Interpretation |
|---|---|---|---|
| gate | 0.500 | 0.500 | Merging perception/consciousness equally; no optimization |
| slot similarity (sim) | ~0 | -0.007 | Slots are orthogonalized, then frozen |
| diversity loss (div) | ~0 | -0.0035 | Loss already minimized at initialization |
| reconstruction loss | ~0.08-0.12 | ~0.08 | Stable; Decoder learns, Memory does not |

The total trainable parameters in SM-v3 are ~26M. With 99,000 images × 1,024 tokens each providing gradient signals, the effective gradient per parameter is still too weak to overcome initialization noise and push parameters away from their starting values. This is not a data problem—it is a gradient scaling problem inherent to the architecture.

## D.2 Why Reconstruction Cannot Train Memory: Five Iterations of SM-v4

While §D.1 exposed the gradient dilution problem in v1-v3 architectures, the question remained: if we could shorten the gradient path to its theoretical minimum, would reconstruction loss suffice to train memory? SM-v4 was designed to answer this with the most favorable conditions possible: a scalar-free, per-slot direct prediction pathway of only 2 MLP layers (gradient path reduced from 8-10 Transformer layers in v1-v3 to 2 feed-forward layers). All experiments used 99,007 unlabeled images, batch_size=16×8 accumulation, lr=5e-5, AdamW optimizer.

Table 3: SM-v4 Iteration Results (30 epochs, H200)

| Variant | Design | direct_loss | grad | sim | Conclusion |
|---|---|---|---|---|---|

| v4.0 | scalar gate + mean_pool + 3-layer MSE | 0.096→0.088 | N/A | -0.005 | gate=0.5 frozen |
|---|---|---|---|---|---|
| v4.1 | no gate, per-slot(256) + MSE ×1 | 0.32→0.30 | 0.004 | -0.005 | gradient reaches but too weak |
| v4.2 | per-slot(256) + MSE ×10 | 2.06→2.02 | 0.003 | -0.011 | 10× pushes sim slightly then plateaus |
| v4.3 | per-slot(256) + DDPM + noisy input | 1.00→0.98 | 0.005 | -0.002 | model exploits noisy shortcut |
| v4.4 | per-slot(512) + DDPM w/o noisy input | 1.00→1.01 | 0.003 | -0.002 | task too difficult for 2-layer MLP |

Key findings:

1. Gradient dilution is not the root cause: v4.1-v4.2 proved that even with a 2-layer MLP gradient path (shortest possible without removing the prediction head entirely), the memory matrix received measurable gradients (grad ≈ 0.003-0.006) yet its content remained effectively static (slot similarity sim moved at most from 0 to -0.011 over 30 epochs). Per-parameter update magnitude ≈ 1e-10 per step, insufficient to overcome initialization.

2. Shortcut learning defeats memory training: When v4.3 provided noisy latent patches as input alongside memory representations, the 2-layer MLP learned to extract noise directly from the noisy input, bypassing memory entirely (direct_loss converged to 1.0—the expected MSE for Gaussian noise, with zero information from memory). This is the same failure mode observed across v1-v3: the model exploits any available shortcut to minimize loss without learning meaningful memory content.

3. The chicken-and-egg deadlock: v4.4 removed all shortcuts (no noisy input, no scalar gate, no mean-pooling), forcing memory to be the sole information source for DDPM denoising. The result was complete failure to converge (direct_loss ≈ 1.0, indicating random-guess prediction). The DDPM task of predicting noise from a 512-dim compressed representation of 16 VQ tokens per slot proved beyond the capacity of a 2-layer MLP. Yet deepening the MLP would reintroduce the gradient dilution problem.

## Appendix E: Local EMA Implementation

```
import torch
import torch.nn.functional as F

# Local EMA Memory Update
# Each image updates only top-k slots per spatial block (sparse routing).
```

```
# No backpropagation -- memory.data is updated in-place.

k_sparse = 8         # top-k slots per spatial block
alpha = 0.01         # update rate (equivalently, 0.99 EMA decay)
B, K, D = write_out.shape   # [batch, 64 spatial blocks, 3584]

# Scaled dot-product similarity: [B, K, K] (block features vs memory slots)
sim = write_out @ memory.data.T / (D ** 0.5)

# Sparse top-k routing: only top-k slots per block receive attention
_, topk_idx = sim.topk(k_sparse, dim=-1)
mask = torch.zeros(B, K, K, device=write_out.device, dtype=write_out.dtype)
mask.scatter_(2, topk_idx, 1.0)
sim = sim.masked_fill(mask == 0, float('-inf'))
attn = F.softmax(sim, dim=-1)              # sparse attention weights

# Weighted local update (gradient-free EMA)
weighted_sum = torch.einsum('bjd,bjk->kd', write_out, attn)  # [K, D]
weights = attn.sum(dim=(0, 1))                              # [K]
for slot in range(K):
    if weights[slot] > 0:
        update = weighted_sum[slot] / weights[slot]
        memory.data[slot] = (1 - alpha) * memory.data[slot] + alpha * update
```